\documentclass[runningheads]{llncs}

\usepackage{eccv}

\usepackage{eccvabbrv}

\usepackage{graphicx}
\usepackage{booktabs}

\usepackage[accsupp]{axessibility}  % Improves PDF readability for those with disabilities.

\usepackage{multirow}  % for \multirow
\usepackage{amssymb}   % for \checkmark
\usepackage{pifont}    % for \xmark (or use alternative)
\usepackage{colortbl}
\usepackage{xcolor}
\newcommand{\xmark}{\ding{55}}
\newcommand{\sparagraph}{

\noindent
\textit}

\usepackage{hyperref}

\usepackage{orcidlink}

\begin{document}

\title{Trajectory-aware Cross-view Geo-localization with Sequential Observations}
% TODO set the abbreviated running title before submission
\titlerunning{Trajectory-aware Cross-view Geo-localization}

\author{Tianyi Gao \and Jiayu Lin \and Danielle Beaulieu \and Nathan Jacobs\orcidlink{0000-0002-4242-8967}}

\authorrunning{T. Gao et al.}

\institute{ Washington University in St. Louis, St. Louis, MO 63130, USA\\ \email{\{t.gao, jiayu.lin, b.danni, jacobsn\}@wustl.edu} }

\maketitle

\begin{abstract}
Cross-view geo-localization matches ground-level observations against geo-tagged satellite imagery. Recent methods show that sequential queries such as video clips yield richer spatiotemporal cues than single images, yet they overlook a complementary sequential modality: route descriptions---which capture the same trajectory at a higher level of abstraction and are often the only input available (e.g., a user directing an autonomous vehicle to a pickup point). To bridge this gap, we introduce SeqGeo-VL, a dataset of $\sim$39K video--text--satellite triplets, and TrajLoc, a unified framework capable of processing both video clips and route descriptions. By leveraging both dense visual and abstract linguistic semantics, TrajLoc enables these modalities to mutually reinforce cross-view matching.
We further propose TrajMod, a lightweight module that conditions query embeddings on trajectory geometry, yielding spatially-aware representations.
Experiments show that TrajLoc achieves substantial gains over state-of-the-art methods on both video and text geo-localization. Code, model weights, and the dataset are released at 
\url{https://humblegamer.github.io/trajloc/}.

\end{abstract}

\section{Introduction}

Urban embodied AI agents, such as autonomous vehicles and quadrupeds, are increasingly deployed for city-wide services~\cite{liu2026urbanverse}. Yet, in crowded urban environments, GPS signals are frequently noisy or completely degraded by tall buildings and vegetation, rendering global localization unreliable~\cite{xia2024text2loc,ye2025cross}. As a result, precise spatial localization based on egocentric visual observations or natural-language descriptions has emerged as a fundamental necessity for agents to navigate safely and cooperate with humans~\cite{anderson2018vision, tian2024loc4plan}, unlocking applications like pedestrian assistance, goods delivery~\cite{xia2024text2loc}, and vehicle pickup~\cite{kolmet2022text2pos}.

Cross-view geo-localization addresses this challenge by matching ground images or textual descriptions against a GPS-tagged satellite image database, casting localization as a retrieval problem~\cite{lin2013cross,zhu2022transgeo,xia2025cross}. Although its granularity is bounded by satellite-patch resolution, it meets the needs of most tasks~\cite{ye2025cross} while offering clear advantages in coverage and storage cost, making it a promising avenue for scalable localization~\cite{ye2025cross}.

\begin{figure}[htbp]
  \centering
  \includegraphics[width=0.9\linewidth]{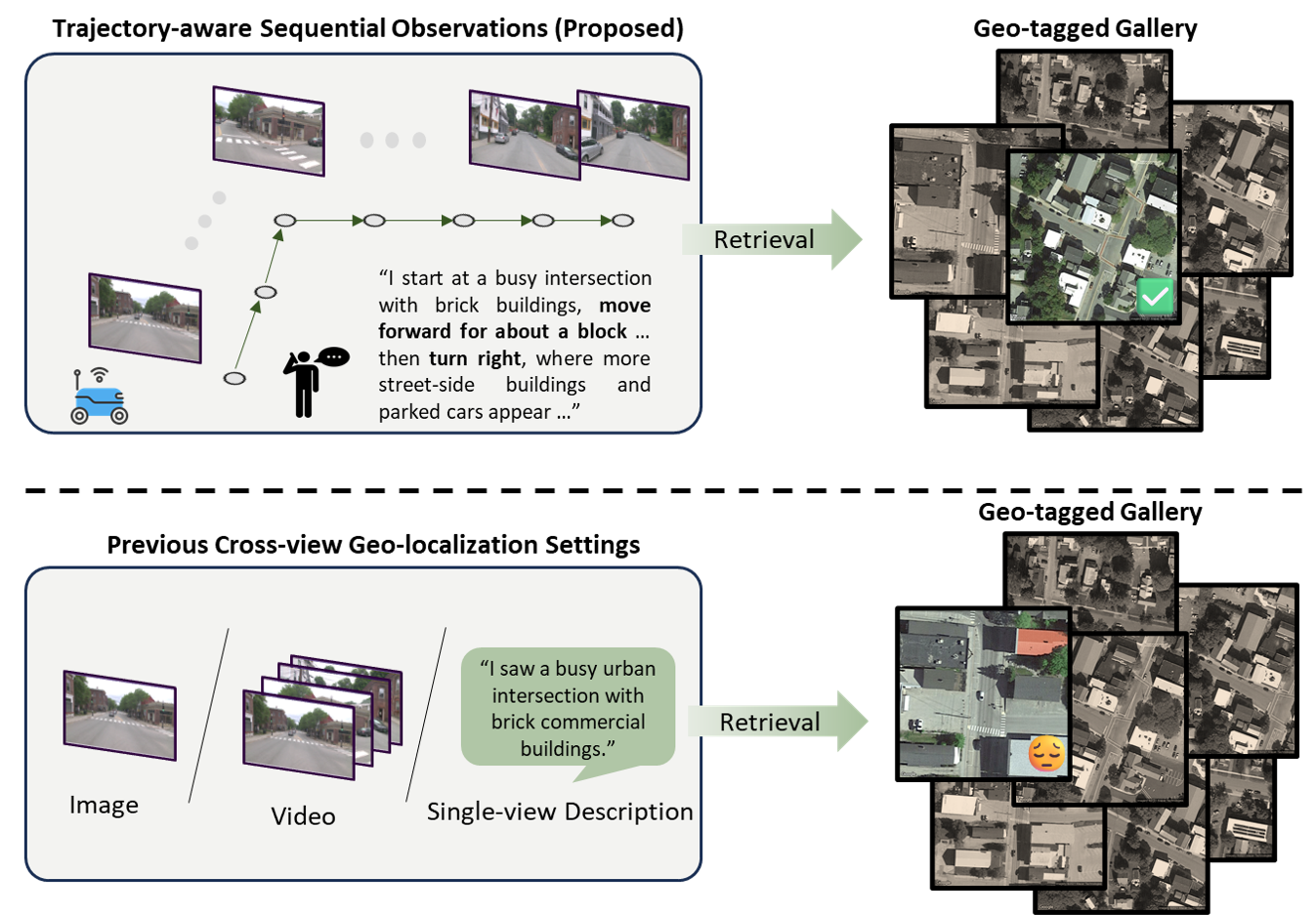}
  \caption{Motivation for our approach, which extends existing cross-view geo-localization methods to sequential route descriptions and explicitly incorporates trajectory information into video/text queries, producing spatially-grounded representations.}
  \label{fig:teaser}
\end{figure}

To overcome perceptual aliasing—where visually identical structures like uniform facades or repeating city blocks cause catastrophic matching ambiguities—recent works~\cite{vyas2022gama, WU2026328, pillai2024garet, shi2022cvlnet} use sequential observations as queries (as illustrated in Fig.~\ref{fig:teaser}). Sequences provide richer cues (motion continuity, changing viewpoints, route-level semantics) that significantly improve retrieval robustness over isolated frames. While this video-to-image retrieval formulation provides dense visual evidence to disambiguate places, it overlooks the textual narrative modality. This limits applications when visual inputs are unavailable~\cite{xiong2024mixed} or during human-robot collaboration, where users naturally rely on abstract language rich in spatiotemporal cues.

However, existing cross-view geo-localization datasets lack route descriptions, let alone video--text--satellite triplets. This restricts flexibility and leaves the potential synergy between route descriptions and videos of a trajectory unexplored.

To address this gap, we develop a scalable annotation pipeline—powered by open-source VLMs and LLMs—that simulates how human users describe routes while moving. Using a progressive strategy, this pipeline combines trajectory-shape priors, frame-level VLM labeling, LLM summarization for concise narratives, and sample-based human verification. We apply this pipeline to the widely used SeqGeo dataset to create \textbf{SeqGeo-VL}, a novel multimodal benchmark comprising 38,863 triplets of trajectory videos, route descriptions, and corresponding geo-referenced satellite images.

Building on SeqGeo-VL, we propose \textbf{TrajLoc}, a unified approach that supports cross-view geo-localization using either video queries or route descriptions. We employ a shared satellite image encoder alongside two separate encoders to process video and route descriptions, respectively. Since directly training modality alignments of varying difficulties can hinder the optimization of the shared satellite image encoder, we adopt a two-stage curriculum learning recipe to maximize the benefits of co-training.

Furthermore, we observe that general CLIP and MLLM embeddings often fail to capture spatial layout, a capability that is crucial for geo-localization since semantically similar objects in different spatial arrangements correspond to distinct locations. We advocate addressing this by better utilizing trajectory information. 
We leverage the fact that modern mobile devices are inherently equipped with IMUs and compasses capable of capturing trajectory geometry alongside video. While often stripped from standard web videos, this information is readily available in active navigation scenarios yet frequently overlooked in cross-view geo-localization methods.
To this end, we introduce \textbf{Trajectory-conditioned Modulation (TrajMod)}, a lightweight plug-and-play module that conditions on trajectory geometry to modulate query embeddings, yielding spatially grounded representations and improving cross-view matching robustness. Our main contributions are:
\begin{itemize}
    \item \textbf{Task and Dataset:} We present a progressive VLM-LLM annotation pipeline to ensure high-quality descriptions and introduce \textbf{SeqGeo-VL}, a multimodal dataset featuring $38,863$ aligned video--text--satellite triplets for evaluating both video- and text-based geo-localization.
    \item \textbf{Unified Approach:} We propose \textbf{TrajLoc}, a strong baseline that handles both video and route description queries via a two-stage curriculum learning strategy, unlocking cross-modal co-training gains.
    \item \textbf{Spatially Grounded Representation:} We introduce \textbf{TrajMod} to inject trajectory geometry into vision-language embeddings. This produces spatially grounded representations that achieve state-of-the-art performance on both cross-view video and text geo-localization tasks.
\end{itemize}

\section{Related Work}

\subsection{Cross-view Geo-localization with Video}
Recent progress in cross-view geo-localization has extended ground-to-aerial image matching to sequential settings with video queries. Many datasets formulate cross-view video geo-localization as a retrieval task, including GaMa~\cite{vyas2022gama} (based on BDD100K~\cite{yu2020bdd100k}), SeqGeo~\cite{zhang2023cross}, and CVLNet~\cite{shi2022cvlnet} (based on KITTI~\cite{geiger2013vision}). While these datasets facilitate learning from sequential observations, they primarily support video-based queries and do not provide natural language descriptions of trajectories, limiting their applicability to scenarios where only route descriptions are available. To bridge this gap, we build upon SeqGeo~\cite{zhang2023cross} and introduce a progressive annotation pipeline to enrich it with trajectory-level textual descriptions, forming a benchmark for cross-view geo-localization with multimodal sequential observations.

Building on these datasets, recent methods focus on mining video context to improve robustness under large viewpoint and appearance changes~\cite{vyas2022gama, zhang2023cross, pillai2024garet, WU2026328}. Focusing on sophisticated aggregation mechanisms for sequential observations, GaMa~\cite{vyas2022gama} utilizes a 3D-CNN, SeqGeo~\cite{zhang2023cross} employs a Temporal Feature Aggregation Module for mixing, GARet~\cite{pillai2024garet} introduces a transformer-based adapter for fusion, and FlexGeo~\cite{WU2026328} uses inter-frame similarity as guidance. Existing approaches rarely exploit explicit trajectory geometry, like per-frame headings or origin--destination layout, to inform representation learning.
Furthermore, existing methods typically assume sequential observations restricted to video format; however, leveraging long-form route descriptions—a scenario of immense practical value for human-robot interaction—remains underexplored.

\subsection{Cross-view Geo-localization with Natural Language Description}
There is abundant work on retrieval tasks between image and text modalities~\cite{wei2026variational, sastry2026probabilistic, xing2026quari}, especially using contrastive learning~\cite{radford2021learning}. However, limited work exists in the domain of cross-view geo-localization via natural language. The goal of this domain is to retrieve the corresponding satellite image from a visually grounded text description. Identifying an image from text that is an exact match to the intended geolocation is an especially challenging task; text often lacks the fine detail required to distinguish one location from another visually similar location. 

The primary prior work pursuing this task, CrossText2Loc~\cite{ye2025cross}, utilizes additional contrastive learning on top of a CLIP backbone to align the embedding spaces of text descriptions and satellite images. They also propose an Expanded Positional Embedding, which accommodates more details in the descriptions. 
As a pioneering contribution, they constructed the CVG-Text dataset and established an effective VLM-based annotation pipeline. However, since their method focuses on describing single isolated street-view images, it does not yet account for the multiple or sequential observations inherent in long-form trajectory descriptions.
Motivated by this, we design a progressive annotation pipeline powered by the synergy of VLM perception and LLM reasoning, applied to video sequences with trajectory metadata.

\subsection{Spatially Grounded Vision-Language Embedding}

Recent research \cite{yuksekgonul2022and} suggests that CLIP behaves as ``bag-of-words'' models, lacking a fundamental comprehension of spatial relationships. Although CLIP exhibits strong zero-shot capabilities across diverse tasks, it consistently struggles with understanding the spatial arrangement of objects within images \cite{wang2025spatialclip}. Wang et al. \cite{wang2025spatialclip} demonstrate that CLIP often fails to distinguish correct spatial descriptions from incorrect ones, sometimes even assigning higher matching scores to captions with erroneous spatial relationships.
Even Multi-modal Large Language Models (MLLMs) with sophisticated decoders struggle to handle relative directions or spatial layouts. The VSI-Bench~\cite{yang2025thinking} highlights that even strong MLLMs with advanced video understanding capabilities cannot accurately map egocentric video frames to allocentric object positions and camera trajectories, an area that remains under active development.

To develop spatially grounded representations, recent studies like SpatialRGPT~\cite{cheng2024spatialrgpt} and SpatialCLIP~\cite{wang2025spatialclip} incorporate depth information with RGB imagery. This approach is designed to bolster the model's grasp of basic spatial properties (e.g., size and length) as well as more intricate relationships, including directionality, spatial adjacency, and perspective transformation.

Towards larger-scale outdoor scenarios, some methods integrate structural data; for instance, UGE \cite{zhang2026urbangraphembeddings} embeds a spatial topology, which is built upon street-view imagery, road networks, and Point of Interest (POI) data, into the VLM using a graph representation. Although this facilitates high-accuracy geo-localization, our approach operates under weaker assumptions regarding available metadata, requiring neither POI data nor road networks.

\section{Problem Formulation and Benchmark}
We first formalize cross-view geo-localization from sequential observations and then describe the construction of our multimodal benchmark for this task.

\subsection{Task Definition}

We study two instances of cross-view geo-localization from sequential observations.
A trajectory $\tau$ is modeled as a sequence of waypoints with relative coordinates $(\Delta x_i, \Delta y_i)$ and headings $h_i$.
Along $\tau$, sequential observations $O_{1:N}$ can be provided as a video $V$ (detailed) or a route description $T$ (abstract).
\begin{equation}
\tau=\{(\Delta x_i,\Delta y_i,h_i)\}_{i=1}^{N}
\quad
O_{1:N}\in\{V, T\}
\end{equation}
Given a geo-tagged satellite gallery $\mathcal{S}=\{S_j\}_{j=1}^{M}$, the goal is to retrieve the matching satellite image:
\begin{equation}
j^{*}=\arg\max_{j\in\{1,\dots,M\}} \ \mathrm{sim}\!\left(\phi_q(O_{1:N},\tau),\, \phi_s(S_j)\right),
\end{equation}
where $\phi_q$ is a video/text query encoder, $\phi_s$ is a satellite encoder, and $\mathrm{sim}(\cdot,\cdot)$ denotes similarity measurement. Incorporating $\tau$ enables a trajectory-aware cross-view geo-localization formulation, while remaining an optional component contingent on trajectory availability. The agent's location is subsequently derived from the geo-tag associated with the retrieved satellite image.

\subsection{Benchmark Dataset}

As shown in Table~\ref{tab:dataset_comparison_compact}, most existing geo-localization datasets rely on single-image queries. While a few explore sequential video queries or single-text queries, none investigate the sequential route description. To address this gap, we extend the SeqGeo dataset~\cite{zhang2023cross} to create SeqGeo-VL, a multimodal geo-localization dataset featuring sequential observations.

To produce high-quality trajectory-level descriptions, we employ a hierarchical annotation pipeline consisting of four stages: 
(1) \textbf{Semantic Extraction}: We utilize Qwen3-VL-8B~\cite{bai2025qwen3} to generate detailed captions for individual frames, capturing fine-grained visual landmarks. 
(2) \textbf{Motion Integration}: We augment these captions with explicit motion primitives (e.g., \textit{turning left}, \textit{proceeding forward}) derived from trajectory metadata. 
(3) \textbf{Structural Summarization}: Qwen3-30B~\cite{yang2025qwen3} distills the comprehensive visual semantics and motion cues into a coherent, trajectory-level narrative. 
This approach ensures that SeqGeo-VL encapsulates both frame-specific details and the global temporal context of the route. (4) \textbf{Quality Control}: Three trained annotators scored a geographically uniform $\sim$1\% subset (400 trajectories) on 3-point object fidelity (OF), 3-point spatiotemporal consistency (ST), and route completeness (RC), with inter-annotator agreement measured by Gwet's AC2. A sample passes if RC is rated complete by a majority and OF$\,+\,$ST$\,>\,3$, giving a pass rate of $91.8\%$. Retrieval performance and word statistics provide further indirect measurement.

\begin{figure}[htbp]
  \centering
  % --- Independent Figure: Word Cloud ---
  \begin{minipage}[t]{0.48\linewidth}
    \centering
    \includegraphics[width=\linewidth]{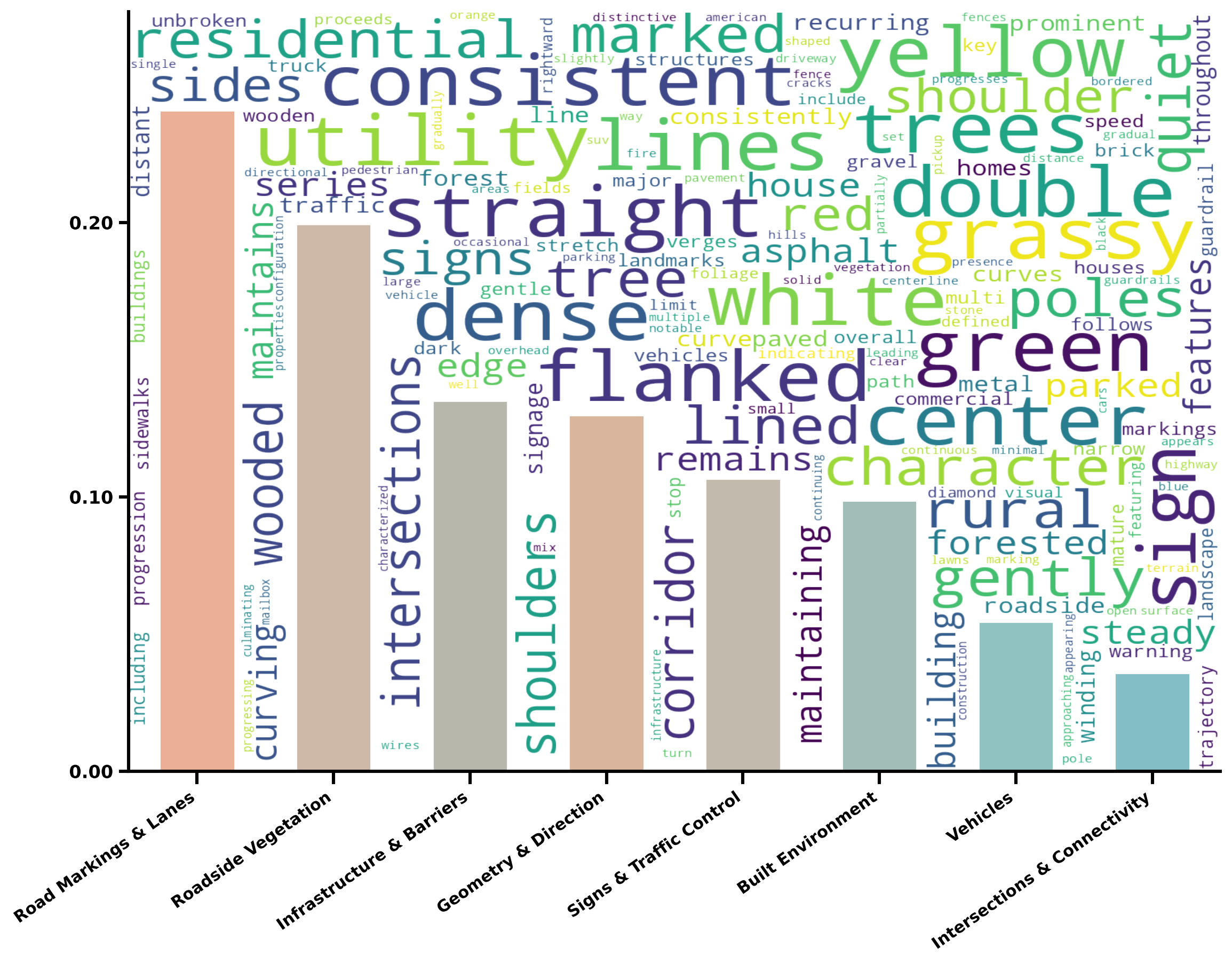}
    \caption{Word cloud and distribution of the extracted keywords within the SeqGeo-VL dataset.}
    \label{fig:wordcloud_hist}
  \end{minipage}
  \hfill
  % --- Independent Figure: Annotation Pipeline ---
  \begin{minipage}[t]{0.48\linewidth}
    \centering
    \includegraphics[width=\linewidth]{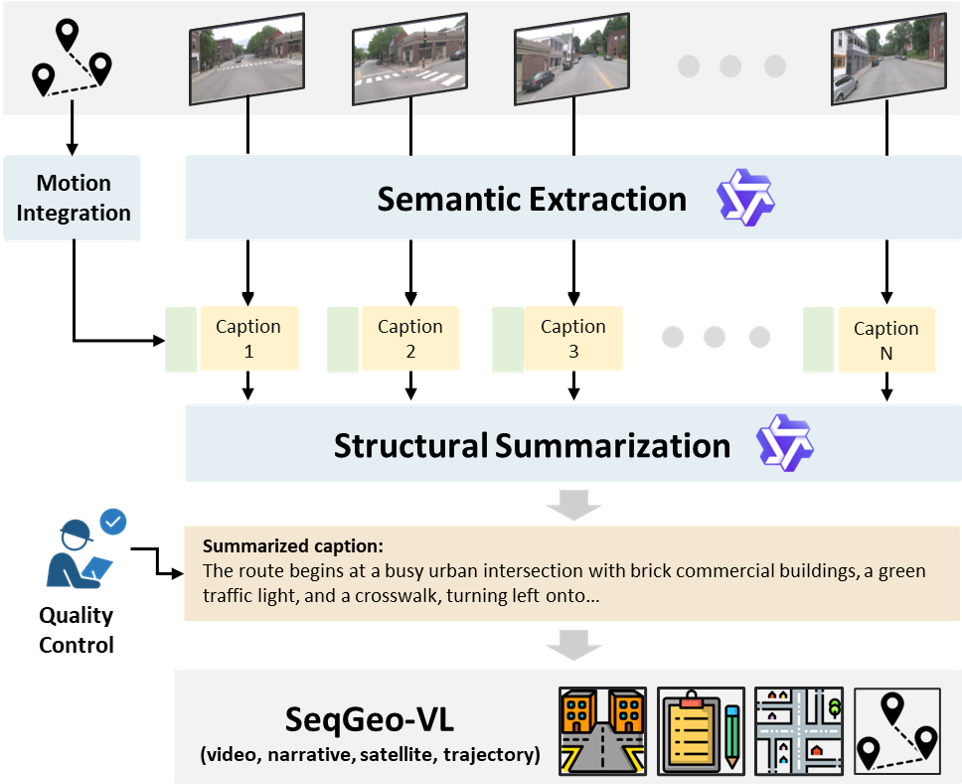}
    \caption{The proposed hierarchical annotation pipeline for SeqGeo-VL.}
    \label{fig:annotation_pipeline}
  \end{minipage}
\end{figure}

The dataset comprises 38,863 triplets consisting of videos, route descriptions, and geo-tagged satellite images at zoom level 20. The distribution of extracted keywords is visualized in Fig.~\ref{fig:wordcloud_hist}, and the entire annotation pipeline is summarized in Fig.~\ref{fig:annotation_pipeline}. Following the protocol of the original SeqGeo~\cite{zhang2023cross}, we split the data into training and testing sets using an 80:20 ratio.

\begin{table*}[htbp]
\centering
\caption{Comparison of cross-view datasets. \textbf{Text} and \textbf{Sequential} indicate whether the dataset supports text modalities and sequential observations, respectively. \textbf{View Type} describes the field of view of the images. \textbf{Qry} and \textbf{Ref} denote the total number of query and reference images. \textbf{Obs} denotes the average number of observations (i.e., frames) associated with each query instance. $^\dagger$ denotes Qry as reported in~\cite{WU2026328}.}

\resizebox{0.8\textwidth}{!}{
\begin{tabular}{l c c c r r r}
\toprule
\textbf{Dataset} 
& \textbf{Text} 
& \textbf{Sequential} 
& \textbf{View Type} 
& \textbf{Qry} 
& \textbf{Ref} 
& \textbf{Obs} \\
\midrule
CVUSA~\cite{workman2015wide}        & \xmark & \xmark & Panoramic    & $\sim$1.5M     & $\sim$1.5M     & $\sim$1 \\
CVACT~\cite{liu2019lending}       & \xmark & \xmark & Panoramic    & 137K    & 137K    & $\sim$1 \\
VIGOR~\cite{zhu2021vigor}       & \xmark & \xmark & Panoramic    & 105K     & 90K    & $\sim$1 \\
CVGlobal~\cite{ye2024cross} & \xmark & \xmark & Panoramic    & 134K    & 134K    & $\sim$1 \\
University-1652~\cite{zheng2020university}  & \xmark & \xmark & Limited FOV  & 5.5K     & 90K     & $\sim$1 \\
\midrule
GeoText-1652~\cite{chu2024towards} 
                         & \checkmark & \xmark & Limited FOV  & 14K     & 90K     & $\sim$1 \\
CVG-Text~\cite{ye2025cross} 
                         & \checkmark & \xmark & Limited FOV  & 30K     & 60K     & $\sim$1 \\
\midrule

KITTI-CVL~\cite{shi2022cvlnet}& \xmark & \checkmark & Limited FOV  & 28K     & 41K     & $\sim$4 \\
GAMa~\cite{vyas2022gama}         & \xmark & \checkmark & Limited FOV  & $\sim$15M$^\dagger$ & 1.9M   & -- \\
SeqGeo~\cite{zhang2023cross}     & \xmark & \checkmark & Limited FOV  & 118K    & 38K     & $\sim$7 \\
SetVL480K~\cite{WU2026328}
                         & \xmark & \xmark & Limited FOV  & 480K    & 16K    & $\sim$4 \\
\midrule
\rowcolor{gray!10}

\textbf{Ours (SeqGeo-VL)} 
                         & \checkmark & \checkmark & Limited FOV & 118K & 38K & $\sim$7 \\
\bottomrule
\end{tabular}
}
\label{tab:dataset_comparison_compact}
\end{table*}

\section{Approach}

\begin{figure}[htbp]
  \centering
  \includegraphics[width=0.9\linewidth]{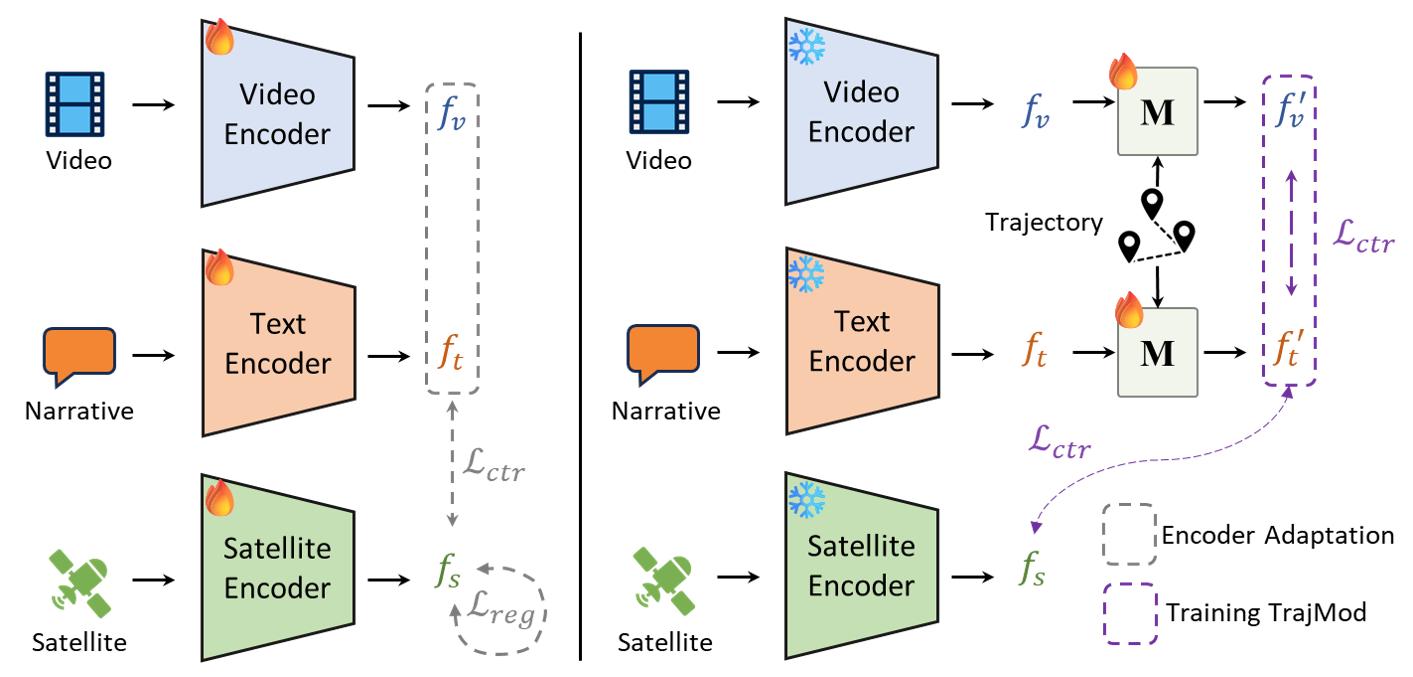}
\caption{Overview of the TrajLoc framework.
  \textbf{Left}: Encoder adaptation with a two-stage curriculum.
  Three CLIP-initialized encoders produce video ($\mathbf{f}_v$),
  text ($\mathbf{f}_t$), and satellite ($\mathbf{f}_s$) embeddings.
  Stage~1 trains the video and satellite encoders with a contrastive
  loss $\mathcal{L}_{\text{ctr}}$; Stage~2 freezes the video encoder
  and aligns the text encoder, with a regularization loss
  $\mathcal{L}_{\text{reg}}$ to prevent satellite-encoder drift.
  \textbf{Right}: All three encoders are frozen, and
  TrajMod (\textbf{M}) conditions the video and text embeddings
  on trajectory geometry to produce spatially grounded representations
  $\mathbf{f}'_v$ and $\mathbf{f}'_t$ for cross-view retrieval.}
  \label{fig:architecture}
\end{figure}

We propose TrajLoc, a unified framework for trajectory-conditioned cross-view geo-localization that supports both video queries and textual route descriptions (Fig.~\ref{fig:architecture}). The architecture comprises three CLIP-initialized encoders---for video, text, and satellite imagery---that are trained with a two-stage curriculum. In stage one, we optimize a video--satellite contrastive objective, providing a strong initialization of the satellite encoder for ground-view semantics. In stage two, we freeze the video encoder and align the text encoder to the satellite encoder with a contrastive loss, applying cosine-similarity regularization to prevent the satellite encoder from drifting.

To facilitate allocentric reasoning, we derive a trajectory geometry embedding from waypoint headings and OD bearing. Conditioned on this, TrajMod predicts scale and shift parameters that modulate the video or text query embeddings, yielding spatially-aware representations for cross-view retrieval.

\subsection{Encoder Adaptation}
We argue that learning text-to-satellite and video-to-satellite retrieval can be mutually beneficial: route descriptions provide abstract, structured semantics, while videos offer dense visual evidence. A unified model also reduces parameters compared to training separate systems. To this end, we instantiate a three-encoder architecture initialized from pretrained CLIP~\cite{radford2021learning}.

\textbf{Video encoder:}
Given a ground-level video $V = \{v_1, v_2, \dots, v_N\}$ consisting of $N$ frames sampled along a trajectory, we independently encode each frame using the CLIP ViT-L/14 image backbone to obtain frame-level features $\{\mathbf{f}_{v_1}, \dots, \mathbf{f}_{v_N}\}$. Following prior video--language models~\cite{Luo2021CLIP4Clip,bolya2025perception}, we aggregate these into a single video embedding via mean pooling: $\mathbf{f}_v = \frac{1}{N}\sum_{i=1}^{N} \mathbf{f}_{v_i}$. Although there are more sophisticated temporal aggregation strategies (e.g. temporal transformers~\cite{zhang2023cross}), we find that mean pooling provides a strong and efficient baseline, keeping the architecture lightweight; improvements can be attributed to the training strategy and trajectory-conditioned modulation rather than encoder complexity.

\textbf{Text encoder: }
Route descriptions in SeqGeo-VL are substantially longer than typical CLIP captions (averaging ${\sim}129$ tokens), exceeding the default CLIP context window of $L_0 = 77$ tokens. Following CrossText2Loc~\cite{ye2025cross}, we initialize the text encoder from OpenAI CLIP and extend its context capacity by linearly interpolating the positional embeddings. 
This preserves the pretrained positional semantics while smoothly extending the context window, enabling the encoder to process long trajectory-level narratives.

\textbf{Satellite encoder: }
We use a CLIP ViT-L/14 image encoder to embed each geo-referenced satellite patch into the shared retrieval space. Because the satellite encoder serves as the common reference for both video and text queries, it plays a central role in the framework. Weights are shared across both training stages and are regularized during the second stage (Sec.~\ref{sec:training}) to prevent catastrophic drift, ensuring that the retrieval space remains coherent for both modalities.

\subsection{Trajectory-conditioned Modulation}
Rather than reducing cross-view geo-localization to standard appearance-based retrieval, we explicitly incorporate trajectory geometry—a widely available yet historically overlooked modality—to spatially ground our representations.

\sparagraph{Trajectory geometry embedding.}
We use two complementary cues (relative to True North): (i) waypoint heading angle $\mathbf{h}=\{h_1,\dots,h_T\}$, capturing local turning patterns, and (ii) the OD bearing angle $\theta_{\text{OD}}$, providing a stable global orientation. These are lightweight to obtain and do not require metric-accurate geometry. To encode angular variation, we apply Fourier features. For any angle $\alpha \in \mathbf{h}\cup\{\theta_{\text{OD}}\}$,
\begin{equation}
\Phi(\alpha)=\big[\sin(2^0\pi\alpha),\cos(2^0\pi\alpha),\dots,\sin(2^{K-1}\pi\alpha),\cos(2^{K-1}\pi\alpha)\big],
\end{equation}
and concatenate the encoded headings and OD bearing along channel dimensions to form a trajectory geometry embedding $\mathbf{z}_{\text{traj}}$.

\sparagraph{Spatially-grounded modulation.}
Inspired by FiLM~\cite{perez2018film}, TrajMod conditions query embeddings on $\mathbf{z}_{\text{traj}}$ via a multi-layer perceptron (MLP) denoted by $\mathrm{MLP}_m$.
Given video and text embeddings $\mathbf{f}_v$ and $\mathbf{f}_t$, we learn modality-specific modulation parameters and transform the features where $\odot$ is element-wise multiplication. We use separate MLPs for video and text to account for the modality gap.
\begin{equation}
\boldsymbol{\gamma}_m,\boldsymbol{\beta}_m=\mathrm{MLP}_m(\mathbf{z}_{\text{traj}}),\quad m\in\{v,t\},\quad
\mathbf{f}'_m=\boldsymbol{\gamma}_m\odot \mathbf{f}_m+\boldsymbol{\beta}_m
\end{equation}

\subsection{Training}
\label{sec:training}

Aligning route descriptions with satellite images is typically harder than aligning videos with satellite images due to the severe modality gap between abstract text and dense overhead pixels. We therefore train TrajLoc with a curriculum to improve both tasks.
In the first stage, we optimize an InfoNCE objective \cite{oord2018representation} $\mathcal{L}_{\text{ctr}}$ to align video and satellite embeddings, yielding a shared space that retains semantics from both views.
In the second stage, we freeze the video encoder and use the satellite encoder as a soft anchor: we align text and satellite embeddings with InfoNCE while regularizing the satellite encoder with a cosine-similarity constraint to mitigate drift from the first stage.
We define a symmetric InfoNCE loss, given a minibatch of $N$ matched pairs $(q_i,k_i)$ (video--satellite or text--satellite), where $\ell(a,b)$ is the standard InfoNCE loss using cosine similarity and a temperature hyperparameter. We also define the regularization loss between the embeddings produced by Stage~1 and Stage~2 satellite encoders $\mathbf{f}_s^{\mathrm{stage1}}$ and $\mathbf{f}_s^{\mathrm{stage2}}$. 
\begin{equation}
\mathcal{L}_{\text{ctr}} = \frac{1}{2N} \sum_{i=1}^{N} \left[ \ell(q_i, k_i) + \ell(k_i, q_i) \right], \quad \mathcal{L}_{\text{reg}} = 1 - \text{sim}(\mathbf{f}_s^{\mathrm{stage1}}, \mathbf{f}_s^{\mathrm{stage2}})
\label{eq:infonce}
\end{equation}

Finally, we freeze the TrajLoc encoders and train TrajMod's two MLPs using the same contrastive objective $\mathcal{L}_{\text{ctr}}$ for video--satellite and text--satellite alignment. Since both modalities are conditioned on the same $\mathbf{z}_{\text{traj}}$, adding a text--video contrastive term serves as a regularizer: it discourages modality-specific noise that is not shared across views and encourages both $\mathbf{f}'_v$ and $\mathbf{f}'_t$ to preserve the common, geometry-grounded factors injected by $\mathbf{z}_{\text{traj}}$.

\section{Experiments}

We evaluate TrajLoc on the SeqGeo-VL dataset, assessing cross-view retrieval from both video and text queries. All experiments use Recall@$K$ (R@$K$) as the primary metric and share a common training recipe, consistent except for a few noted exceptions, to ensure a fair comparison.

\subsection{Experimental Setup and Implementation Details}

We evaluate our framework on two tasks: cross-view text geo-localization~\cite{ye2025cross}, which retrieves a geo-referenced satellite image from a natural language route description, and cross-view video geo-localization~\cite{zhang2023cross, pillai2024garet}, which matches ground-level videos to geo-referenced satellite imagery.
 We adopt the pretrained CLIP ViT-L/14 as the backbone for all three encoders. The model is optimized using Adam with an initial learning rate of $1 \times 10^{-5}$. Video- and text-alignment with satellite imagery are each trained for 40 epochs with a total batch size of 24, while TrajMod is trained for 20 epochs with a batch size of 128. Full implementation details of the compared models are provided in the supplementary material.

\begin{figure}[tbp]
    \centering
    \includegraphics[width=\linewidth]{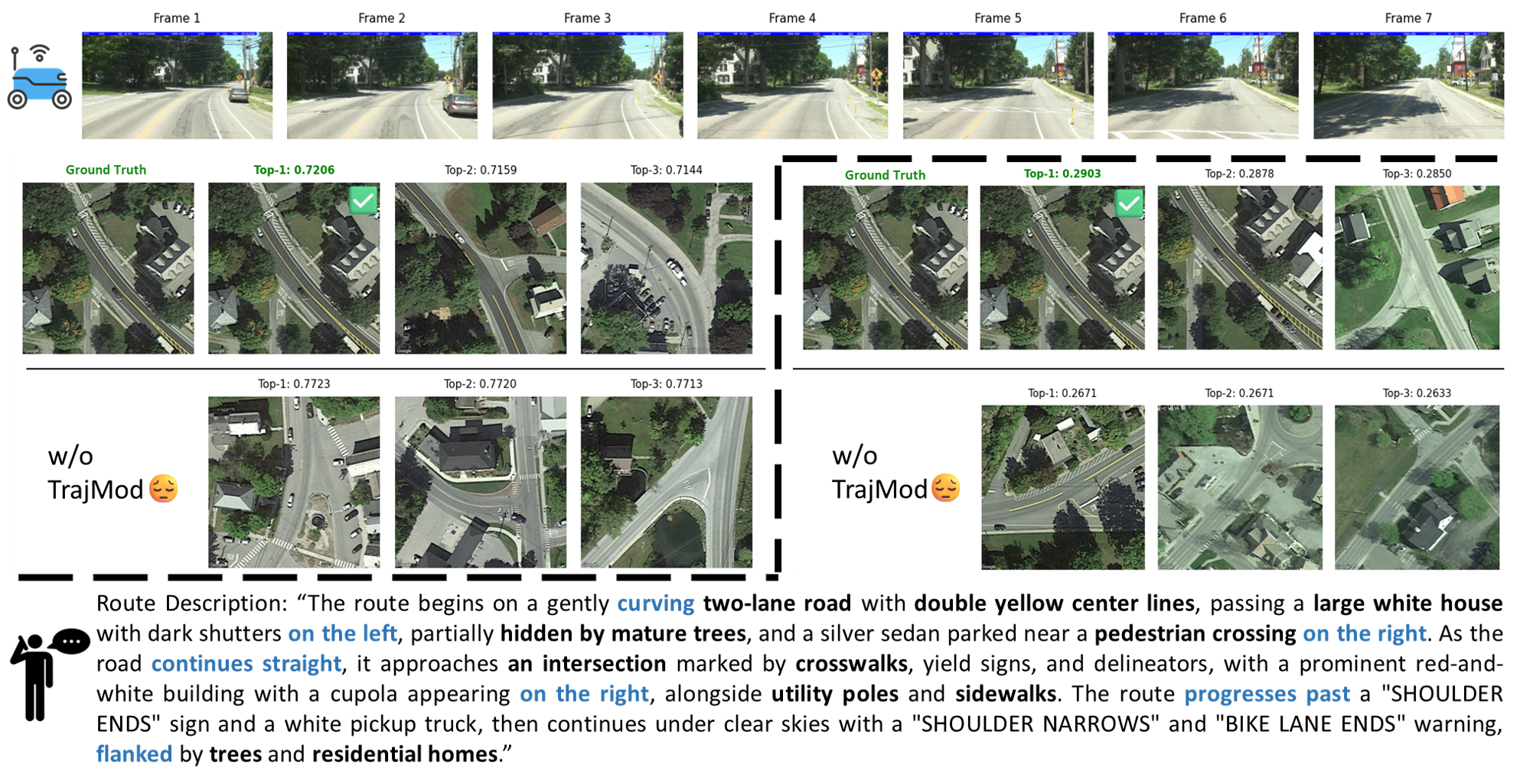}
    \caption{Cross-view geo-localization with and without TrajMod. \textcolor{RoyalBlue}{\textbf{Blue}} highlights the spatial cues, while \textbf{bold} denotes the observed objects. As observed in Top-3 candidates, TrajMod enables more layout-aligned retrieval results.}
    \label{fig:qualitative}
\end{figure}

\subsection{Cross-view Video Geo-localization Experiments}
For cross-view video geo-localization, we compare our method with recent state-of-the-art (SOTA) models, including FlexGeo~\cite{WU2026328}, GARet~\cite{pillai2024garet}, and SeqGeo~\cite{zhang2023cross}. To mitigate the impact of backbone selection, we re-implemented SeqGeo using a CLIP ViT-L/14 architecture (denoted as SeqGeo$^\dagger$). Notably, our video encoding method employs simple mean pooling~\cite{Luo2021CLIP4Clip} for efficiency, whereas SeqGeo utilizes multiple transformer layers for feature aggregation.

As shown in Table~\ref{tab:video_quantitative}, TrajLoc achieves superior performance on cross-view video geo-localization, reaching an R@1 of 12.09\%. Upgrading the SeqGeo backbone from VGG16 to CLIP ViT-L/14 improves its R@1 from 1.80\% to 8.14\%; however, TrajLoc still outperforms it by a significant margin using the same ViT-L/14 backbone, even while operating with a lower computational budget.

We further evaluate Qwen3-VL-Embedding, as its MLLM-based pre-training on large-scale multimodal data provides strong video understanding. We apply LoRA fine-tuning to equip it with cross-view reasoning capabilities, since end-to-end adaptation of MLLMs is substantially more resource-intensive under practical constraints. However, it still underperforms our approach on this challenging task, as general-purpose VLM embeddings lack the spatially grounded characteristics essential for cross-view matching.

\begin{table}[htbp]
\centering
\caption{Comparison of cross-view video geo-localization results. $^*$denotes LoRA fine-tuning~\cite{hu2022lora}; $^{\dagger}$denotes reimplementation using the original code. Otherwise, results are excerpted from the original papers.}
\label{tab:video_quantitative}
\addtolength{\tabcolsep}{3pt}
\resizebox{0.9\textwidth}{!}{
\begin{tabular}{ll|cccc}
\toprule
\textbf{Method} & \textbf{Backbone} & \textbf{R@1} & \textbf{R@5} & \textbf{R@10} & \textbf{R@1\%} \\
\midrule
SeqGeo~\cite{zhang2023cross}  & VGG16            & 1.80  & 6.45  & 10.36 & 34.38 \\
SeqGeo$^\dagger$~\cite{zhang2023cross}  & ViT-L/14 & 8.14  & 24.42 & 34.02 & 66.17 \\
GARet~\cite{pillai2024garet}   & DeiT-m         & 3.34  & 11.19 & 17.18 & 44.39 \\
FlexGeo~\cite{WU2026328} & ConvNext-B     & 3.51  & 14.22 & 20.77 & 48.53 \\
Qwen3-VL-Embedding$^*$~\cite{bai2025qwen3} & Qwen3-VL-2B & 5.44  & 19.21 & 28.15 & 61.16 \\
\midrule
\rowcolor{gray!10}
TrajLoc (Ours) & ViT-L/14       & \textbf{12.09} & \textbf{35.77} & \textbf{47.68} & \textbf{80.21} \\
\bottomrule
\end{tabular}
}
\end{table}

\subsection{Cross-view Text Geo-localization Experiments}
For cross-view text geo-localization, we compare our method with recent state-of-the-art (SOTA) models, including CrossText2Loc~\cite{ye2025cross} and CLIP, EVA2-CLIP, and SigLIP as also analyzed by Ye et al.~\cite{ye2025cross}. Since cross-view text geo-localization is a new task~\cite{ye2025cross}, limited prior work exists for comparison. We also include recent general multimodal encoders, Perception Encoder and Qwen3-VL-Embedding. Most baselines share a similar text encoder design of extending the context window by \emph{linearly interpolating} positional embeddings.

Table~\ref{tab:text_quantititive} shows TrajLoc consistently achieves the best performance (R@1=2.52\%, R@1\%=45.48\%), surpassing the strongest baselines by a large margin: $2.6\times$ higher R@1 than CrossText2Loc (ViT-L/14@336) and $1.7\times$ higher R@1\% than EVA2-CLIP (ViT-L/14@336).
Notably, scaling the backbone (B$\rightarrow$L$\rightarrow$SO400M) or increasing input resolution (@336/@384) yields only modest gains for CLIP-style methods, suggesting that the bottleneck lies not in backbone capacity but in the spatial reasoning required to bridge egocentric route descriptions and allocentric satellite layouts.
Overall, the results indicate that current VLM/MLLM embeddings still struggle with egocentric-to-allocentric spatial reasoning, motivating TrajLoc’s trajectory-conditioned design to explicitly bridge narrative semantics with route geometry for more precise cross-view retrieval.

\begin{table}[htbp]
\centering
\caption{Comparison of cross-view text geo-localization results. $^*$denotes LoRA fine-tuning~\cite{hu2022lora}; otherwise, all models are fully fine-tuned on SeqGeo-VL.}
\label{tab:text_quantititive}
\resizebox{0.8\linewidth}{!}{
\begin{tabular}{ll|cccc}
\toprule
\textbf{Method} & \textbf{Backbone} & \textbf{R@1} & \textbf{R@5} & \textbf{R@10} & \textbf{R@1\%} \\
\midrule
CLIP~\cite{radford2021learning}               & ViT-B/16      & 0.53 & 1.58 & 3.33  & 15.94 \\
CLIP~\cite{radford2021learning}               & ViT-L/14      & 0.80 & 3.32 & 6.01  & 22.99 \\
EVA2-CLIP~\cite{fang2024eva}          & ViT-B/16      & 0.63 & 2.20 & 3.98  & 17.59 \\
EVA2-CLIP~\cite{fang2024eva}          & ViT-L/14@336  & 0.89 & 3.78 & 6.70  & 26.99 \\
SigLIP~\cite{zhai2023sigmoid}             & ViT-B/16      & 0.54 & 2.21 & 4.07  & 18.22 \\
SigLIP~\cite{zhai2023sigmoid}             & ViT-L/16@384  & 0.85 & 3.13 & 5.24  & 21.33 \\
SigLIP~\cite{zhai2023sigmoid}             & ViT-SO400M/14        & 0.80 & 2.92 & 5.24  & 21.86 \\
Perception Enc.~\cite{bolya2025perception}    & ViT-L/14@336  & 0.66 & 2.08 & 3.80  & 15.17 \\
Qwen3-VL-Embed.$^*$~\cite{bai2025qwen3}    & Qwen3-VL-2B   & 0.93 & 3.38 & 5.55  & 20.91 \\
CrossText2Loc~\cite{ye2025cross}      & ViT-L/14      & 0.84 & 3.29 & 5.73  & 22.29 \\
CrossText2Loc~\cite{ye2025cross}      & ViT-L/14@336  & 0.98 & 4.08 & 7.08  & 25.45 \\
\midrule
\rowcolor{gray!10}
TrajLoc (Ours)     & ViT-L/14      & \textbf{2.52} & \textbf{9.82} & \textbf{16.11} & \textbf{45.48} \\
\bottomrule
\end{tabular}
}
\end{table}

\subsection{Ablation study}
\begin{figure}[htbp]
    \centering
    \includegraphics[width=\linewidth]{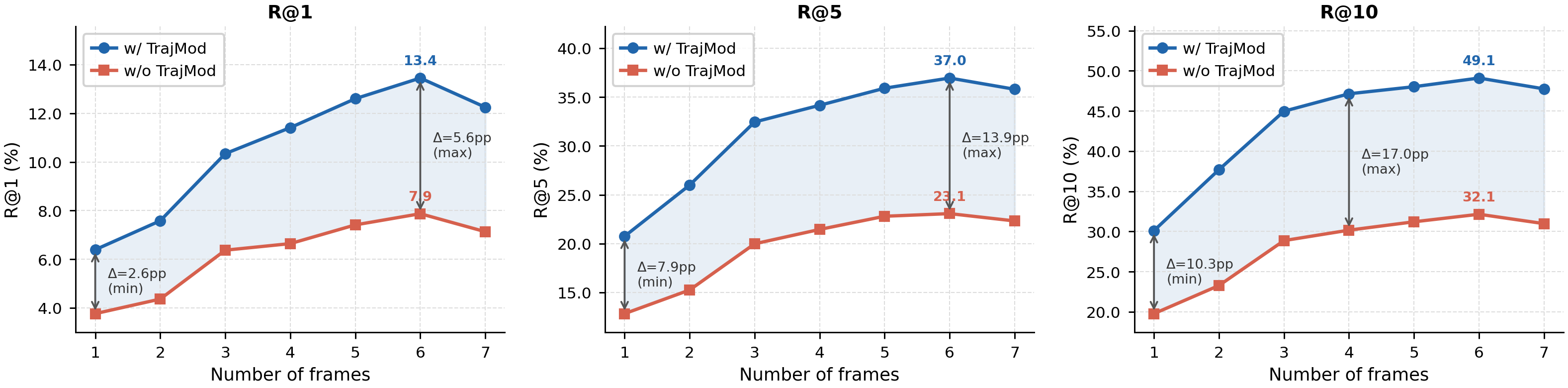}
    \caption{Video geo-localization performance with varying frame number and TrajMod.}
    \label{fig:vid2sat_compare}
\end{figure}

\sparagraph{Impact of Temporal Length and Trajectory Geometry.}
To investigate the interplay between sequence length and trajectory conditioning, we evaluate video-to-satellite retrieval using varying numbers of input frames, with and without TrajMod. As shown in Fig.~\ref{fig:vid2sat_compare}, performance generally improves across all metrics (R@1, R@5, and R@10) as the number of frames grows from 1 to 6, validating our core motivation that extended temporal context provides richer cues for cross-view geo-localization compared to single-frame observations. Moreover, the trajectory-conditioned variant (\textit{w/ TrajMod}) consistently outperforms the baseline (\textit{w/o TrajMod}), with the performance gap steadily widening as sequences lengthen---e.g., the R@1 margin ($\Delta$) grows from 2.6\% at 1 frame to 5.6\% at 6 frames. 
We attribute this to the fact that longer sequential observations inevitably suffer from stronger visual aliasing and feature smoothing. Rather than relying on temporal modeling or heuristic frame selection, TrajMod modulates semantic features conditioned on the geometry of the viewpoint trajectory, capturing the spatial layout of observed semantics.
The qualitative results are shown in Fig.~\ref{fig:qualitative}.

\begin{figure}[htbp]
    \centering
    \includegraphics[width=\linewidth]{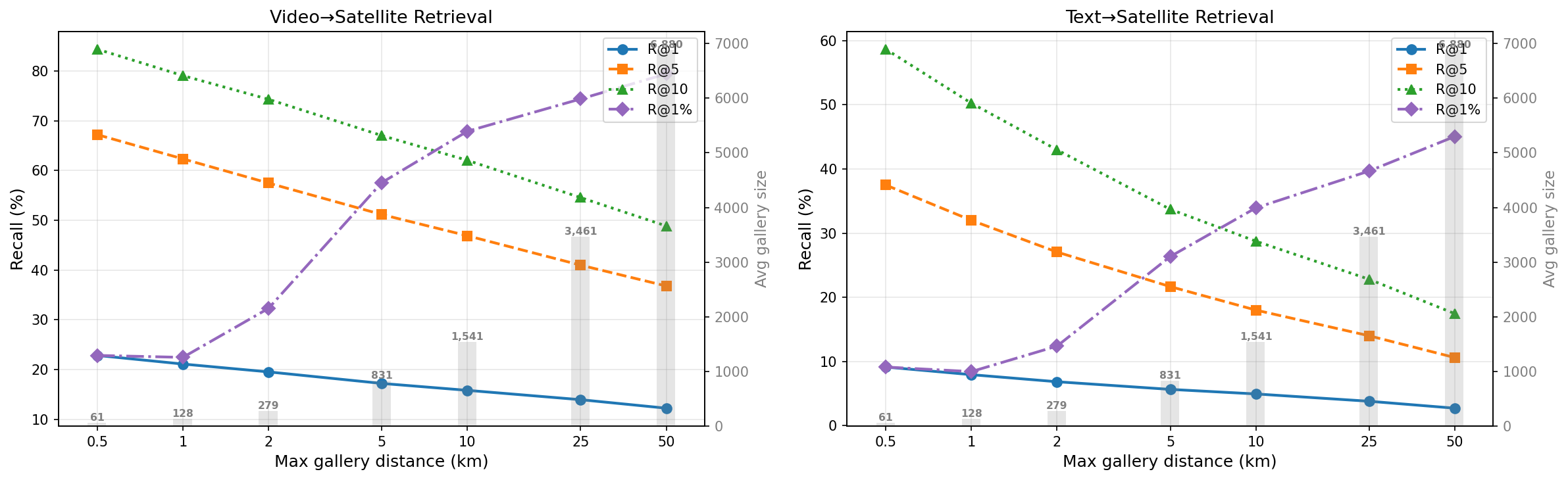}
    \caption{Performance with different gallery sizes given a location prior.}
    \label{fig:vary_gallery}
\end{figure}

\sparagraph{Location Prior Context.}
In practical applications, users or mobile robots often operate with a location prior, such as knowing their rough geographic district when GPS signals are severely obscured. To simulate these conditions, we evaluate retrieval performance under varying gallery sizes, effectively narrowing the search space based on prior knowledge. As shown in Fig.~\ref{fig:vary_gallery}, restricting the candidate pool substantially improves the absolute retrieval accuracy and demonstrates the practical utility of our explicitly grounded embeddings.

\sparagraph{Synergy between Co-training and TrajMod.} 
We investigate the mutual benefits of jointly training cross-view geo-localization with visual or linguistic sequential observations, and how TrajMod amplifies this synergy. As reported in Table~\ref{tab:ablation}, without TrajMod, simply co-training the two modalities yields only marginal changes over independent training (e.g., Video R@1 improves slightly from 6.87\% to 7.27\%, and Text R@1 from 0.80\% to 0.98\%). 

However, when TrajMod is introduced, the gains from co-training are significantly amplified. For the video-to-satellite task, co-training with the text modality boosts R@1 by a remarkable 2.40\% (from 9.69\% to 12.09\%). Similarly, for text-to-satellite, co-training with video improves R@1 from 1.90\% to 2.52\%. We attribute this synergy to TrajMod’s explicit geometric grounding. By anchoring visual observations and textual narratives to the same trajectory geometry, TrajMod establishes a shared spatial representation that facilitates cross-modal knowledge transfer, allowing visual details and route-level semantics to mutually reinforce each other.

\begin{table}[htbp]
\centering
\caption{Ablation study on co-training components and TrajMod.}
\label{tab:ablation}
\addtolength{\tabcolsep}{2pt}
\resizebox{0.9\textwidth}{!}{%
\begin{tabular}{@{}l cccc cccc@{}}
\toprule
& \multicolumn{4}{c}{\textbf{Video Geo-localization}} & \multicolumn{4}{c}{\textbf{Text Geo-localization}} \\
\cmidrule(lr){2-5} \cmidrule(l){6-9}
 & R@1 & R@5 & R@10 & R@1\% & R@1 & R@5 & R@10 & R@1\% \\
\midrule
\rowcolor{gray!10}
\multicolumn{9}{@{}l}{\small\textit{(a) TrajMod component ablation}} \\

Full model & \textbf{12.09} & \textbf{35.77} & \textbf{47.68} & \textbf{80.21} & \textbf{2.52} & \textbf{9.82} & \textbf{16.11} & 45.48 \\
\quad \textit{w/o} Txt2Vid contrastive loss & 10.23 & 30.57 & 41.21 & 74.20 & 2.17 & 8.49 & 14.19 & 42.22 \\
\quad \textit{w/o} TrajMod & 7.27 & 22.13 & 30.92 & 63.47 & 0.98 & 4.16 & 7.21 & 26.15 \\
\midrule
\rowcolor{gray!10}
\multicolumn{9}{@{}l}{\small\textit{(b) Co-training ablation} (w/ TrajMod)} \\
Full model  & \textbf{12.09} & \textbf{35.77} & \textbf{47.68} & \textbf{80.21} & \textbf{2.52} & \textbf{9.82} & \textbf{16.11} & 45.48 \\
\quad Vid2Sat only & 9.69 & 29.75 & 40.30 & 73.60 & -- & -- & -- & -- \\
\quad Txt2Sat only & -- & -- & -- & -- & 1.90 & 9.12 & 15.57 & \textbf{47.25} \\
\midrule
\rowcolor{gray!10}
\multicolumn{9}{@{}l}{\small\textit{(c) Co-training ablation} (w/o TrajMod)} \\
Full model & 7.27 & 22.13 & 30.92 & 63.47 & 0.98 & 4.16 & 7.21 & 26.15 \\
\quad Vid2Sat only & 6.87 & 22.50 & 31.60 & 64.09 & -- & -- & -- & -- \\
\quad Txt2Sat only & -- & -- & -- & -- & 0.80 & 3.32 & 6.01 & 22.99 \\
\quad \textit{w/o} Curriculum learning & 5.01 & 18.28 & 27.25 & 61.81 & 0.86 & 3.86 & 6.52 & 24.40 \\
\bottomrule
\end{tabular}%
}
\end{table}

\sparagraph{Utilizing Trajectory Geometry.}
We compare two methods for incorporating the same trajectory information: text prompting for MLLM-based embeddings (e.g., Qwen3-VL) and TrajMod for our CLIP-style co-trained architectures (Table~\ref{tab:traj_conditioning}). While adding trajectory details via prompts yields a modest improvement (+1.34\% for R@1\%), TrajMod provides a substantial gain of +19.33\% for R@1\%. This demonstrates that explicit geometric modulation is significantly more effective than text-based instructions for spatial reasoning in cross-view retrieval.

\sparagraph{Co-sequenced Multimodal Query.}
We further examine the performance of our unified approach when both video and text queries are available.
Linearly blending video and text query embeddings ($\alpha=0.4$) consistently outperforms the stronger single-modality baseline across all frame counts, with R@10 gains ranging from $+9.49$ at 1 frame down to $+4.75$ at 7 frames; the complete per-frame results are reported in the supplementary material. This indicates text and video provide complementary route cues, with text especially valuable when video is sparse, missing, or corrupted---a favorable storage/encoding trade-off.

\subsection{Discussion on SeqGeo-VL}
To evaluate the necessity of sequential observations and spatiotemporal cues, we compare SeqGeo-VL against two variants (Table~\ref{tab:text_quality_retrieval}) with CrossText2Loc~\cite{ye2025cross}. 

First, \textit{Single-view Description} uses only the VLM-generated description of the final waypoint. This yields the lowest performance (10.63\% R@1\%), indicating a single-frame observation lacks the context for unambiguous matching. 

Second, to verify the importance of temporal order, we design \textit{Sequential-view Objects}. To construct this variant, we extract perceived entities and nouns from the route descriptions and order them chronologically to maintain their temporal order of appearance. Despite its minimal length (26.3 words), it significantly outperforms the single-view baseline (17.41\% vs. 10.63\% R@1\%), validating that sequential observations provide critical localization cues. 

Finally, the full SeqGeo-VL achieves the best results (22.29\% R@1\%). The performance gap between the object-only variant and the full dataset demonstrates that relational context (i.e., spatiotemporal cues) is essential for constructing spatially-aware language representations. SeqGeo-VL also exhibits a lower average similarity, indicating more diverse textual descriptions.

\begin{table}[htbp]
    \centering
    \begin{minipage}{0.58\textwidth}
        \centering
        \caption{Text quality and retrieval results.} 
        \label{tab:text_quality_retrieval}
        \resizebox{\textwidth}{!}{%
        \begin{tabular}{@{}l cc cc@{}} 
            \toprule
            \multirow{2}{*}{\textbf{Dataset}} 
            & \multicolumn{2}{c}{\textbf{Text quality}} 
            & \multicolumn{2}{c}{\textbf{Retrieval}} \\
            \cmidrule(lr){2-3} \cmidrule(l){4-5} 
            & \textbf{Word Len.} & \textbf{Avg Sim.} & \textbf{R@1} & \textbf{R@1\%} \\
            \midrule
            Single-view caption & 50.9 & 0.270 & 0.36 & 10.63 \\
            Sequential-view objects & 26.3 & 0.281 & 0.58 & 17.41 \\
            \rowcolor{gray!10}
            SeqGeo-VL & \textbf{101.3} & \textbf{0.247} & \textbf{0.84} & \textbf{22.29} \\
            \bottomrule
        \end{tabular}}
    \end{minipage}
    \hfill 
    \begin{minipage}{0.38\textwidth}
        \centering
        \caption{Comparison of trajectory conditioning mechanisms.}
        \label{tab:traj_conditioning}
        \resizebox{\textwidth}{!}{%
        \begin{tabular}{@{}l c c@{}}
            \toprule
            \textbf{Method} & \textbf{R@1\%} & \textbf{$\Delta$} \\
            \midrule
            Qwen3-VL-Embed.~\cite{bai2025qwen3} & 20.91 & - \\
            \rowcolor{gray!10}
            + Prompt & 22.25 & +1.34 \\
            \midrule
            Our CLIP-style baseline & 26.15 & - \\
            \rowcolor{gray!10}
            + TrajMod & \textbf{45.48} & \textbf{+19.33} \\
            \bottomrule
        \end{tabular}}
    \end{minipage}
\end{table}

\section{Conclusion}

We extend route-level cross-view geo-localization beyond video queries to natural-language route descriptions. We introduce SeqGeo-VL, a multimodal benchmark of video-text-satellite triplets for this task; TrajLoc, a unified approach for video and text queries; and TrajMod, a lightweight module that incorporates explicit trajectory geometry for feature modulation. TrajLoc achieves state-of-the-art performance on cross-view geo-localization from both videos and route descriptions. These results demonstrate the value of jointly modeling sequential visual and linguistic observations, and highlight the role of trajectory geometry in complementing vision-language representations for cross-view spatial reasoning.

\section*{Acknowledgements}

This research used the TGI RAILS advanced compute and data resource which is supported by the National Science Foundation (award OAC-2232860) and the Taylor Geospatial Institute.

\bibliographystyle{splncs04}
\bibliography{main}

@String(CVPR  = {IEEE Conf. Comput. Vis. Pattern Recog.})

@String(ICLR  = {Int. Conf. Learn. Represent.})

@String(AAAI  = {AAAI})

@String(CVPR  = {CVPR})

@String(ICLR  = {ICLR})

@inproceedings{workman2015wide,
  title={Wide-area image geolocalization with aerial reference imagery},
  author={Workman, Scott and Souvenir, Richard and Jacobs, Nathan},
  booktitle={Proceedings of the IEEE International Conference on Computer Vision},
  pages={3961--3969},
  year={2015}
}

@inproceedings{zhu2021vigor,
  title={Vigor: Cross-view image geo-localization beyond one-to-one retrieval},
  author={Zhu, Sijie and Yang, Taojiannan and Chen, Chen},
  booktitle={Proceedings of the IEEE/CVF Conference on Computer Vision and Pattern Recognition},
  pages={3640--3649},
  year={2021}
}

@inproceedings{ye2024cross,
  title={Cross-view image geo-localization with Panorama-BEV Co-Retrieval Network},
  author={Ye, Junyan and Lv, Zhutao and Li, Weijia and Yu, Jinhua and Yang, Haote and Zhong, Huaping and He, Conghui},
  booktitle={European Conference on Computer Vision},
  pages={74--90},
  year={2024},
  organization={Springer}
}

@inproceedings{anderson2018vision,
  title={Vision-and-language navigation: Interpreting visually-grounded navigation instructions in real environments},
  author={Anderson, Peter and Wu, Qi and Teney, Damien and Bruce, Jake and Johnson, Mark and S{\"u}nderhauf, Niko and Reid, Ian and Gould, Stephen and Van Den Hengel, Anton},
  booktitle={Proceedings of the IEEE conference on computer vision and pattern recognition},
  pages={3674--3683},
  year={2018}
}

@inproceedings{zheng2020university,
  title={University-1652: A multi-view multi-source benchmark for drone-based geo-localization},
  author={Zheng, Zhedong and Wei, Yunchao and Yang, Yi},
  booktitle={Proceedings of the 28th ACM international conference on Multimedia},
  pages={1395--1403},
  year={2020}
}

@inproceedings{zhang2023cross,
  title={Cross-view image sequence geo-localization},
  author={Zhang, Xiaohan and Sultani, Waqas and Wshah, Safwan},
  booktitle={Proceedings of the IEEE/CVF Winter Conference on Applications of Computer Vision},
  pages={2914--2923},
  year={2023}
}

@inproceedings{vyas2022gama,
  title={Gama: Cross-view video geo-localization},
  author={Vyas, Shruti and Chen, Chen and Shah, Mubarak},
  booktitle={European Conference on Computer Vision},
  pages={440--456},
  year={2022},
  organization={Springer}
}

@article{hu2022lora,
  title={Lora: Low-rank adaptation of large language models.},
  author={Hu, Edward J and Shen, Yelong and Wallis, Phillip and Allen-Zhu, Zeyuan and Li, Yuanzhi and Wang, Shean and Wang, Liang and Chen, Weizhu and others},
  journal={Iclr},
  volume={1},
  number={2},
  pages={3},
  year={2022}
}

@article{geiger2013vision,
  title={Vision meets robotics: The kitti dataset},
  author={Geiger, Andreas and Lenz, Philip and Stiller, Christoph and Urtasun, Raquel},
  journal={The international journal of robotics research},
  volume={32},
  number={11},
  pages={1231--1237},
  year={2013},
  publisher={Sage Publications Sage UK: London, England}
}

@Article{Luo2021CLIP4Clip,
  author  = {Huaishao Luo and Lei Ji and Ming Zhong and Yang Chen and Wen Lei and Nan Duan and Tianrui Li},
  title   = {{CLIP4Clip}: An Empirical Study of CLIP for End to End Video Clip Retrieval},
  journal = {arXiv preprint arXiv:2104.08860},
  year    = {2021},
}

@inproceedings{ye2025cross,
  title={Where am i? cross-view geo-localization with natural language descriptions},
  author={Ye, Junyan and Lin, Honglin and Ou, Leyan and Chen, Dairong and Wang, Zihao and Zhu, Qi and He, Conghui and Li, Weijia},
  booktitle={Proceedings of the IEEE/CVF International Conference on Computer Vision},
  pages={5890--5900},
  year={2025}
}

@inproceedings{liu2019lending,
  title={Lending orientation to neural networks for cross-view geo-localization},
  author={Liu, Liu and Li, Hongdong},
  booktitle={Proceedings of the IEEE/CVF conference on computer vision and pattern recognition},
  pages={5624--5633},
  year={2019}
}

@article{WU2026328,
title = {Set-CVGL: A new perspective on cross-view geo-localization with unordered ground-view image sets},
journal = {ISPRS Journal of Photogrammetry and Remote Sensing},
volume = {233},
pages = {328-345},
year = {2026},
issn = {0924-2716},
doi = {https://doi.org/10.1016/j.isprsjprs.2026.01.037},
url = {https://www.sciencedirect.com/science/article/pii/S0924271626000456},
author = {Qiong Wu and Panwang Xia and Lei Yu and Yi Liu and Mingtao Xiong and Liheng Zhong and Jingdong Chen and Ming Yang and Yongjun Zhang and Yi Wan}
}

@inproceedings{radford2021learning,
  title={Learning transferable visual models from natural language supervision},
  author={Radford, Alec and Kim, Jong Wook and Hallacy, Chris and Ramesh, Aditya and Goh, Gabriel and Agarwal, Sandhini and Sastry, Girish and Askell, Amanda and Mishkin, Pamela and Clark, Jack and others},
  booktitle={International conference on machine learning},
  pages={8748--8763},
  year={2021},
  organization={PmLR}
}

@inproceedings{kolmet2022text2pos,
  title={Text2pos: Text-to-point-cloud cross-modal localization},
  author={Kolmet, Manuel and Zhou, Qunjie and O{\v{s}}ep, Aljo{\v{s}}a and Leal-Taix{\'e}, Laura},
  booktitle={Proceedings of the IEEE/CVF conference on computer vision and pattern recognition},
  pages={6687--6696},
  year={2022}
}

@inproceedings{zhai2023sigmoid,
  title={Sigmoid loss for language image pre-training},
  author={Zhai, Xiaohua and Mustafa, Basil and Kolesnikov, Alexander and Beyer, Lucas},
  booktitle={Proceedings of the IEEE/CVF international conference on computer vision},
  pages={11975--11986},
  year={2023}
}

@article{fang2024eva,
  title={Eva-02: A visual representation for neon genesis},
  author={Fang, Yuxin and Sun, Quan and Wang, Xinggang and Huang, Tiejun and Wang, Xinlong and Cao, Yue},
  journal={Image and Vision Computing},
  volume={149},
  pages={105171},
  year={2024},
  publisher={Elsevier}
}

@article{zhang2026urbangraphembeddings,
  title={UrbanGraphEmbeddings: Learning and Evaluating Spatially Grounded Multimodal Embeddings for Urban Science},
  author={Zhang, Jie and Yu, Xingtong and Fang, Yuan and Stouffs, Rudi and Trivic, Zdravko},
  journal={arXiv preprint arXiv:2602.08342},
  year={2026}
}

@inproceedings{pillai2024garet,
  title={Garet: cross-view video geolocalization with adapters and auto-regressive transformers},
  author={Pillai, Manu S and Rizve, Mamshad Nayeem and Shah, Mubarak},
  booktitle={European Conference on Computer Vision},
  pages={466--483},
  year={2024},
  organization={Springer}
}

@article{yuksekgonul2022and,
  title={When and why vision-language models behave like bags-of-words, and what to do about it?},
  author={Yuksekgonul, Mert and Bianchi, Federico and Kalluri, Pratyusha and Jurafsky, Dan and Zou, James},
  journal={arXiv preprint arXiv:2210.01936},
  year={2022}
}

@inproceedings{yang2025thinking,
  title={Thinking in space: How multimodal large language models see, remember, and recall spaces},
  author={Yang, Jihan and Yang, Shusheng and Gupta, Anjali W and Han, Rilyn and Fei-Fei, Li and Xie, Saining},
  booktitle={Proceedings of the Computer Vision and Pattern Recognition Conference},
  pages={10632--10643},
  year={2025}
}

@inproceedings{wang2025spatialclip,
  title={Spatialclip: Learning 3d-aware image representations from spatially discriminative language},
  author={Wang, Zehan and Zhou, Sashuai and He, Shaoxuan and Huang, Haifeng and Yang, Lihe and Zhang, Ziang and Cheng, Xize and Ji, Shengpeng and Jin, Tao and Zhao, Hengshuang and others},
  booktitle={Proceedings of the Computer Vision and Pattern Recognition Conference},
  pages={29656--29666},
  year={2025}
}

@article{cheng2024spatialrgpt,
  title={Spatialrgpt: Grounded spatial reasoning in vision-language models},
  author={Cheng, An-Chieh and Yin, Hongxu and Fu, Yang and Guo, Qiushan and Yang, Ruihan and Kautz, Jan and Wang, Xiaolong and Liu, Sifei},
  journal={Advances in Neural Information Processing Systems},
  volume={37},
  pages={135062--135093},
  year={2024}
}

@article{bolya2025perception,
  title={Perception encoder: The best visual embeddings are not at the output of the network},
  author={Bolya, Daniel and Huang, Po-Yao and Sun, Peize and Cho, Jang Hyun and Madotto, Andrea and Wei, Chen and Ma, Tengyu and Zhi, Jiale and Rajasegaran, Jathushan and Rasheed, Hanoona and others},
  journal={arXiv preprint arXiv:2504.13181},
  year={2025}
}

@inproceedings{shi2022cvlnet,
  title={Cvlnet: Cross-view semantic correspondence learning for video-based camera localization},
  author={Shi, Yujiao and Yu, Xin and Wang, Shan and Li, Hongdong},
  booktitle={Asian Conference on Computer Vision},
  pages={123--141},
  year={2022},
  organization={Springer}
}

@inproceedings{perez2018film,
  title={Film: Visual reasoning with a general conditioning layer},
  author={Perez, Ethan and Strub, Florian and De Vries, Harm and Dumoulin, Vincent and Courville, Aaron},
  booktitle={Proceedings of the AAAI conference on artificial intelligence},
  volume={32},
  number={1},
  year={2018}
}

@inproceedings{chu2024towards,
  title={Towards natural language-guided drones: GeoText-1652 benchmark with spatial relation matching},
  author={Chu, Meng and Zheng, Zhedong and Ji, Wei and Wang, Tingyu and Chua, Tat-Seng},
  booktitle={European Conference on Computer Vision},
  pages={213--231},
  year={2024},
  organization={Springer}
}

@inproceedings{yu2020bdd100k,
  title={Bdd100k: A diverse driving dataset for heterogeneous multitask learning},
  author={Yu, Fisher and Chen, Haofeng and Wang, Xin and Xian, Wenqi and Chen, Yingying and Liu, Fangchen and Madhavan, Vashisht and Darrell, Trevor},
  booktitle={Proceedings of the IEEE/CVF conference on computer vision and pattern recognition},
  pages={2636--2645},
  year={2020}
}

@article{bai2025qwen3,
  title={Qwen3-vl technical report},
  author={Bai, Shuai and Cai, Yuxuan and Chen, Ruizhe and Chen, Keqin and Chen, Xionghui and Cheng, Zesen and Deng, Lianghao and Ding, Wei and Gao, Chang and Ge, Chunjiang and others},
  journal={arXiv preprint arXiv:2511.21631},
  year={2025}
}

@inproceedings{xia2024text2loc,
  title={Text2loc: 3d point cloud localization from natural language},
  author={Xia, Yan and Shi, Letian and Ding, Zifeng and Henriques, Joao F and Cremers, Daniel},
  booktitle={Proceedings of the IEEE/CVF conference on computer vision and pattern recognition},
  pages={14958--14967},
  year={2024}
}

@article{yang2025qwen3,
  title={Qwen3 technical report},
  author={Yang, An and Li, Anfeng and Yang, Baosong and Zhang, Beichen and Hui, Binyuan and Zheng, Bo and Yu, Bowen and Gao, Chang and Huang, Chengen and Lv, Chenxu and others},
  journal={arXiv preprint arXiv:2505.09388},
  year={2025}
}

@article{oord2018representation,
  title={Representation learning with contrastive predictive coding},
  author={Oord, Aaron van den and Li, Yazhe and Vinyals, Oriol},
  journal={arXiv preprint arXiv:1807.03748},
  year={2018}
}

@inproceedings{tian2024loc4plan,
  title={Loc4plan: Locating before planning for outdoor vision and language navigation},
  author={Tian, Huilin and Meng, Jingke and Zheng, Wei-Shi and Li, Yuan-Ming and Yan, Junkai and Zhang, Yunong},
  booktitle={Proceedings of the 32nd ACM International Conference on Multimedia},
  pages={4073--4081},
  year={2024}
}

@inproceedings{liu2026urbanverse,
  title={UrbanVerse: Scaling Urban Simulation by Watching City-Tour Videos},
  author={Mingxuan Liu and Honglin He and Elisa Ricci and Wayne Wu and Bolei Zhou},
  booktitle={The Fourteenth International Conference on Learning Representations},
  year={2026},
}

@inproceedings{lin2013cross,
  title={Cross-view image geolocalization},
  author={Lin, Tsung-Yi and Belongie, Serge and Hays, James},
  booktitle={Proceedings of the IEEE Conference on Computer Vision and Pattern Recognition},
  pages={891--898},
  year={2013}
}

@inproceedings{zhu2022transgeo,
  title={Transgeo: Transformer is all you need for cross-view image geo-localization},
  author={Zhu, Sijie and Shah, Mubarak and Chen, Chen},
  booktitle={Proceedings of the IEEE/CVF Conference on Computer Vision and Pattern Recognition},
  pages={1162--1171},
  year={2022}
}

@article{xia2025cross,
  title={Cross-view geo-localization with panoramic street-view and VHR satellite imagery in decentrality settings},
  author={Xia, Panwang and Yu, Lei and Wan, Yi and Wu, Qiong and Chen, Peiqi and Zhong, Liheng and Yao, Yongxiang and Wei, Dong and Liu, Xinyi and Ru, Lixiang and others},
  journal={ISPRS Journal of Photogrammetry and Remote Sensing},
  volume={227},
  pages={1--11},
  year={2025},
  publisher={Elsevier}
}

@article{xiong2024mixed,
  title={Mixed-view panorama synthesis using geospatially guided diffusion},
  author={Xiong, Zhexiao and Xing, Xin and Workman, Scott and Khanal, Subash and Jacobs, Nathan},
  journal={Transactions on Machine Learning Research}
}

@inproceedings{
wei2026variational,
title={Variational Adapter for Cross-modal Similarity Representation},
author={WenZhang Wei and Zhipeng Gui and Dehua Peng and Tiandi Ye and Huayi Wu},
booktitle={Forty-third International Conference on Machine Learning},
year={2026},
url={https://openreview.net/forum?id=BkZ40FNKlL}
}

@InProceedings{sastry2026probabilistic,
    author    = {Sastry, Srikumar and Choudhury, Anustup and Madala, Pavani and Su, Guan-Ming},
    title     = {Probabilistic Multimodal Learning with Bayesian Disagreements},
    booktitle = {Proceedings of the IEEE/CVF Conference on Computer Vision and Pattern Recognition (CVPR) Workshops},
    month     = {June},
    year      = {2026},
    pages     = {7546-7555}
}

@inproceedings{xing2026quari,
 author = {Xing, Eric and Stylianou, Abby and Pless, Robert and Jacobs, Nathan},
 booktitle = {Advances in Neural Information Processing Systems},
 editor = {D. Belgrave and C. Zhang and H. Lin and R. Pascanu and P. Koniusz and M. Ghassemi and N. Chen},
 pages = {62890--62911},
 publisher = {Curran Associates, Inc.},
 title = {QuARI: Query Adaptive Retrieval Improvement},
 url = {https://proceedings.neurips.cc/paper_files/paper/2025/file/5ae6634cb92414dc34d24e863e8d0809-Paper-Conference.pdf},
 volume = {38},
 year = {2025}
}
\end{document}